\documentclass[10pt,twocolumn,letterpaper]{article}

\usepackage{iccv}
\usepackage{times}
\usepackage{epsfig}
\usepackage{graphicx}
\usepackage{amsmath}
\usepackage{amssymb}
\usepackage{booktabs}
\usepackage{multirow}
\usepackage{lib}
\usepackage[font=small,labelfont=bf,skip=5pt]{caption}

\usepackage[pagebackref=true,breaklinks=true,letterpaper=true,colorlinks,bookmarks=false]{hyperref}

\iccvfinalcopy 

\def\iccvPaperID{0751} 

\ificcvfinal\fi
\begin{document}

\title{Actions and Attributes from Wholes and Parts}

\author{Georgia Gkioxari\\
UC Berkeley\\
{\tt\small gkioxari@eecs.berkeley.edu}
\and
Ross Girshick\\
Microsoft Research\\
{\tt\small rbg@microsoft.com}
\and
Jitendra Malik\\
UC Berkeley \\
{\tt\small malik@eecs.berkeley.edu}
}

\maketitle

\begin{abstract}
   We investigate the importance of parts for the tasks of action and attribute classification. We develop a part-based approach by leveraging convolutional network features inspired by recent advances in computer vision. Our part detectors are a deep version of poselets and capture parts of the human body under a distinct set of poses. For the tasks of action and attribute classification, we train holistic convolutional neural networks and show that adding parts leads to top-performing results for both tasks. In addition, we demonstrate the effectiveness of our approach when we replace an oracle person detector, as is the default in the current evaluation protocol for both tasks, with a state-of-the-art person detection system.
 \end{abstract}


\section{Introduction}
\seclabel{intro}

For the tasks of human attribute and action classification, it is difficult to infer from the recent literature if part-based modeling is essential or, to the contrary, obsolete. Consider action classification. Here, the method from Oquab \etal \cite{Oquab14} uses a holistic CNN classifier that outperforms part-based approaches \cite{MajiActionCVPR11,CVPR11_0254}. Turning to attribute classification, Zhang \etal 's CNN-based PANDA system \cite{panda} shows that parts bring dramatic improvements over a holistic CNN model. How should we interpret these results? We aim to bring clarity by presenting a single approach for both tasks that shows consistent results.

We develop a part-based system, leveraging convolutional network features, and apply it to attribute and action classification. For both tasks, we find that a properly trained holistic model matches current approaches, while parts contribute further. Using deep CNNs we establish new top-performing results on the standard PASCAL human attribute and action classification benchmarks.

\figref{fig1} gives an outline of our approach. We compute CNN features on a set of bounding boxes associated with the instance to classify. One of these bounding boxes corresponds to the whole instance and is either provided by an oracle or comes from a person detector. The other bounding boxes (three in our implementation) come from poselet-like part detectors.

\begin{figure}[t]
\begin{center}
  \includegraphics[width=0.95\linewidth]{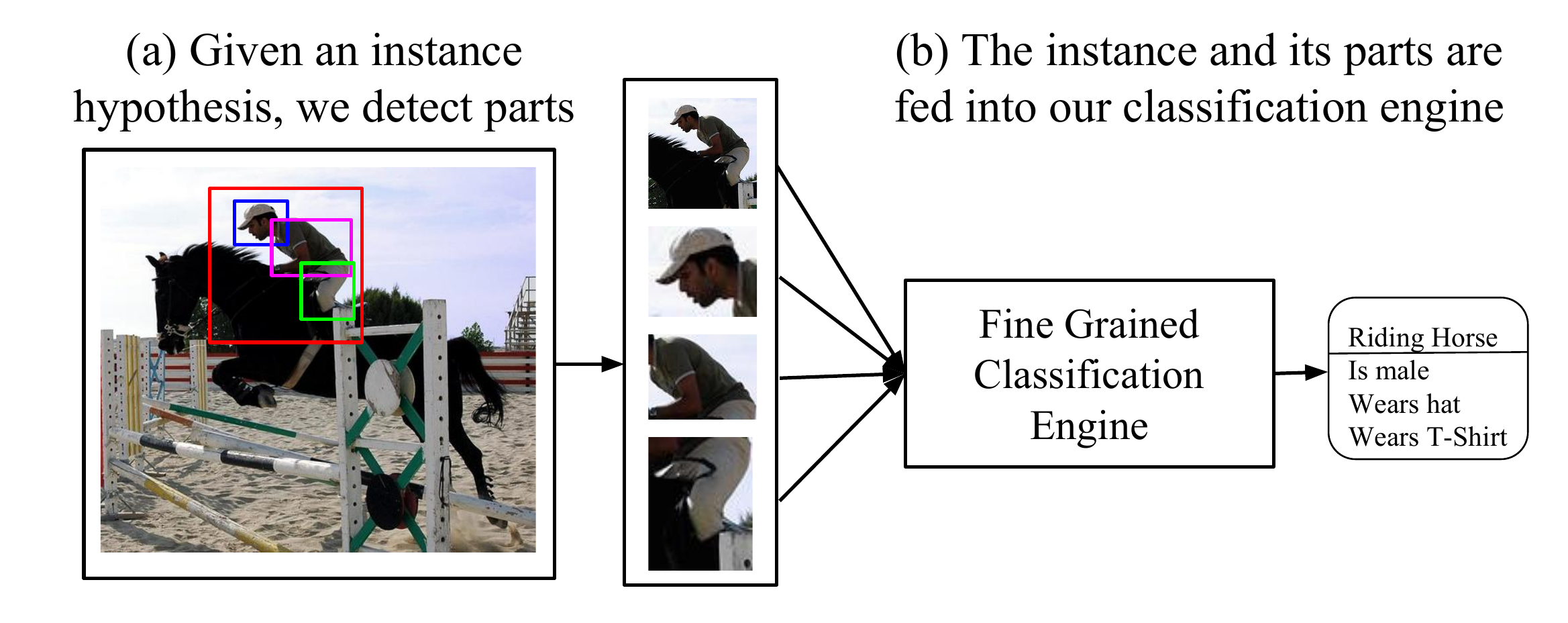}
\end{center}
\caption{Schematic overview of our overall approach. (a) Given an R-CNN person detection (red box), we detect parts using a novel, deep version of poselets (\secref{parts}). (b) The detected whole-person and part bouding boxes are input into a fine-grained classification engine to produce predictions for actions and attributes (\secref{tasks}).}
   \figlabel{fig1}
\end{figure}

Our part detectors are a novel ``deep'' version of poselets. We define three human body parts (head, torso, and legs) and cluster the keypoints of each part into several distinct poselets. Traditional poselets \cite{BourdevPoseletsECCV10,BourdevMalikICCV09} would then operate as sliding-window detectors on top of low-level gradient orientation features, such as HOG \cite{Dalal05}. Instead, we train a sliding-window detector for each poselet on top of a deep feature pyramid, using the implementation of \cite{deeppyramid}. Unlike HOG-based poselets, our parts are capable of firing on difficult to detect structures, such as sitting versus standing legs. Also, unlike recent deep parts based on bottom-up regions \cite{zhang14finegrained}, our sliding-window parts can span useful, but inhomogeneous regions, that are unlikely to group together through a bottom-up process (\eg, bare arms and a t-shrit).

Another important aspect of our approach is task-specific CNN fine-tuning. We show that a fine-tuned holistic model (\ie, no parts) is capable of matching the attribute classification performance of the part-based PANDA system \cite{panda}. Then, when we add parts our system outperforms PANDA. This result indicates that PANDA's dramatic improvement from parts comes primarily from the weak holistic classifier baseline used in their work, rather than from the parts themselves. While we also observe an improvement from adding parts, our marginal gain over the holistic model is smaller, and the gain becomes even smaller as our network becomes deeper. This observation suggests a possible trend: as more powerful convolutional network architectures are engineered, the marginal gain from explicit parts may vanish. 

As a final contribution, we show that our system can operate ``without training wheels.''  In the standard evaluation protocol for benchmarking attributes and actions \cite{BourdevAttributesICCV11,PASCAL-ijcv}, an oracle provides a perfect bounding box for each test instance. While this was a reasonable ``cheat'' a couple of years ago, it is worth revisiting. Due to recent substantial advances in detection performance, we believe it is time to drop the oracle bounding box at test time. We show, for the first time, experiments doing just this; we replace ground-truth bounding boxes with person detections from a state-of-the-art R-CNN person detector \cite{girshick2014rcnn}. Doing so only results in a modest drop in performance compared to the traditional oracle setting.

\section{Related Work}
\seclabel{related}

\noindent
\textbf{Low-level image features}. 
Part-based approaches using low-level features have been successful for a variety of computer vision tasks. DPMs \cite{lsvm-pami} capture different aspects of an object using mixture components and deforable parts, leading to good performance on object detection and attribute classification \cite{DPD}. Similarly, poselets \cite{BourdevPoseletsECCV10,BourdevMalikICCV09,armlets2013,kposelets,MajiActionCVPR11,wang2011learning} are an ensemble of models which capture parts of an object under different viewpoints and have been used for object detection, action and attribute classification and pose estimation. Pictorial structures and its variants \cite{Felzenszwalb05,Fischler73,yang2012articulated} explicitly model parts of objects and their geometric relationship in order to accurately predict their location.

\noindent
\textbf{Convolutional network features}.
Turning away from hand-designed feature representations, convolutional networks (CNNs) have shown remarkable results on computer vision tasks, such as digit recognition \cite{lecun-89e} and more recently image classification \cite{krizhevsky2012imagenet, vgg}. Girshick \etal \cite{girshick2014rcnn} show that a holistic CNN-based approach performs significantly better than previous methods on object detection. They classify region proposals using a CNN fine-tuned on object boxes. Even though their design has no explicit part or component structure, it is able to detect objects under a wide variety of appearance and occlusion patterns.

\noindent
\textbf{Hybrid feature approaches.}
Even more recently, a number of methods incorporate HOG-based parts into deep models, showing significant improvements. Zhang \etal \cite{panda} use HOG-poselet activations and train CNNs, one for each poselet type, for the task of attribute classification. They show a large improvement on the task compared to HOG-based approaches. However, their approach includes a number of suboptimal choices. They use pretrained HOG poselets to detect parts and they train a ``shallow'' CNN (by today's standards) from scratch using a relatively small dataset of 25k images. We train poselet-like part detectors on a much richer feature representation than HOG, derived from the pool5 layer of \cite{krizhevsky2012imagenet}. Indeed, \cite{girshick2014rcnn,deeppyramid} show an impressive jump in object detection performance using pool5 instead of HOG. In addition, the task-specific CNN that we use for action or attribute classification shares the architecture of \cite{krizhevsky2012imagenet, vgg} and is initialized by pre-training on the large ImageNet-1k dataset prior to task-specific fine-tuning.

In the same vein, Branson \etal \cite{branson2014} tackle the problem of bird species categorization by first detecting bird parts with a HOG-DPM and then extracting CNN features from the aligned parts. They experimentally show the superiority of CNN-based features to hand-crafted representations. However, they work from relatively weak HOG-DPM part detections, using CNNs solely for classification purposes. Switching to the person category, HOG-DPM does not generate accurate part/keypoint predictions as shown by \cite{kposelets}, and thus cannot be regarded as a source for well aligned body parts.

\noindent
\textbf{Deep parts.}
Zhang \etal \cite{zhang14finegrained} introduce part-based R-CNNs for the task of bird species classification. They discover parts of birds from region proposals and combine them for classification. They gain from using parts and also from fine-tuning a CNN for the task starting from ImageNet weights. However, region proposals are not guaranteed to produce parts. Most techniques, such as \cite{UijlingsIJCV2013}, are designed to generate candidate regions that contain whole objects based on bottom-up cues. While this approach works for birds, it may fail in general as parts can be defined arbitrarily in an object and need not be of distinct color and texture with regard to the rest of the object. Our sliding-window parts provide a more general solution. Indeed, we find that the recall of selective search regions for our parts is 15.6\% lower than our sliding-window parts across parts at 50\% intersection-over-union.

Tompson \etal \cite{tompson2014} and Chen and Yuille \cite{chen2014} train keypoint specific part detectors, in a CNN framework, for human body pose estimation and show significant improvement compared to \cite{toshev2014}. Their models assume that all parts are visible or self-occluded, which is reasonable for the datasets they show results on.
The data for our task contain significantly more clutter, truncation, and occlusion and so our system is designed to handle missing parts.

Bourdev \etal \cite{deepposelets} introduce a form of deep poselets by training a network with a cross entropy loss. Their system uses a hybrid approach which first uses HOG poselets to bootstrap the collection of training data. They substitute deep poselets in the poselet detection pipeline \cite{BourdevPoseletsECCV10} to create person hypothesis. Their network is smaller than \cite{krizhevsky2012imagenet} and they train it from scratch without hard negative mining. They show a marginal improvement over R-CNN for person detection, after feeding their hypothesis through R-CNN for rescoring and bounding box regression. Our parts look very much like poselets, since they capture parts of a pose. However, we cluster the space of poses instead of relying on random selection and train our models using a state-of-the-art network \cite{krizhevsky2012imagenet} with hard negative mining.

\section{Deep part detectors}
\seclabel{parts}

\figref{DeepParts} schematically outlines the design of our deep part detectors, which can be viewed as a multi-scale fully convolutional network. The first stage produces a feature pyramid by convolving the levels of the gaussian pyramid of the input image with a 5-layer CNN, similar to Girshick \etal \cite{deeppyramid} for training DeepPyramid DPMs. The second stage outputs a pyramid of part scores by convolving the feature pyramid with the part models.

\begin{figure*}[t]
\begin{center}
  \includegraphics[width=0.9\linewidth]{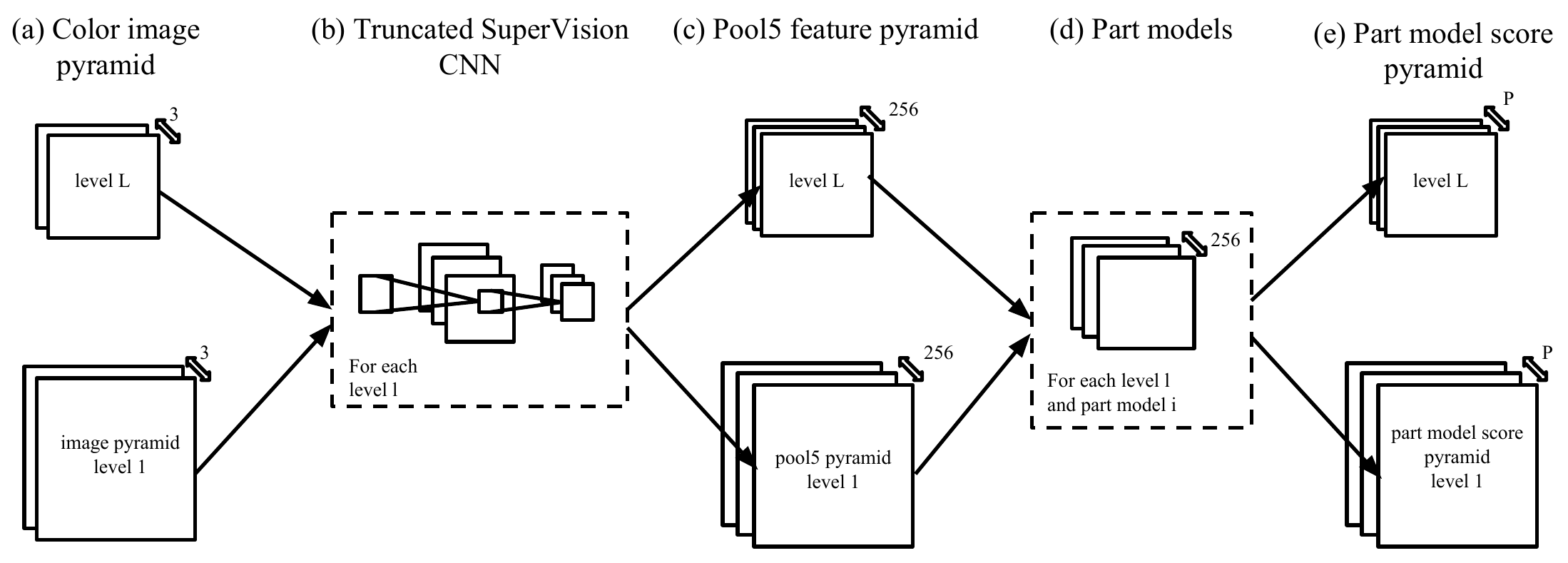}
\end{center}
   \caption{Schematic overview of our part detectors. (a) A gaussian pyramid is build from an input image. (b) Each level of the pyramid is fed into a truncated SuperVision CNN. (c) The output is a pyramid of pool5 feature maps. (d) Each level of the feature pyramid is convolved with the part models. (e) The output is a pyramid of part model scores}
   \figlabel{DeepParts}
\end{figure*}

\subsection{Feature pyramid}

Feature pyramids allow for object and part detections at multiple scales while the corresponding models are designed at a single scale. This is one of the oldest ``tricks" in computer vision and has been implemented by sliding-window object detection approaches throughout the years \cite{BourdevPoseletsECCV10,lsvm-pami,sermanet-iclr-14,Vaillant94anoriginal}.

Given an input image, the construction of the feature pyramid starts by creating the gaussian pyramid for the image for a fixed number of scales and subsequently extracting features from each scale. For feature extraction, we use a CNN and more precisely, we use a variant of the single-scale network proposed by Krizhevsky \etal \cite{krizhevsky2012imagenet}. 
More details can be found in \cite{deeppyramid}. Their software is publicly available, we draw on their implementation.


\subsection{Part models}

We design models to capture parts of the human body under a particular viewpoint and pose. Ideally, part models should be (a) pose-sensitive, \ie produce strong activations on examples of similar pose and viewpoint, (b) inclusive, \ie cover all the examples in the training set, and (c) discriminative, \ie score higher on the object than on the background. To achieve all the above properties, we build part models by clustering the keypoint configurations of all the examples in the training set and train linear SVMs on pool5 features with hard negative mining.

\subsubsection{Designing parts}

We model the human body with three high-level parts: the head, the torso and the legs. Even though the pose of the parts is tied with the global pose of the person, each one has it own degrees of freedom. In addition, there is a large, yet not infinite due to the kinematic constraints of the human body, number of possible part combinations that cover the space of possible human poses.

We design parts defined by the three body areas, head ($H$), torso ($T$) and legs ($L$). Assume $t \in \{H,T,L\}$ and $K_t^{(i)}$ the set of 2D keypoints of the $i$-th training example corresponding to part $t$. The keypoints correspond to predefined landmarks of the human body. Specifically, $K_H=\{\textit{Eyes, Nose, Shoulders}\}$,  $K_T=\{\textit{Shoulders, Hips}\}$ and for $K_L=\{\textit{Hips, Knees, Ankles}\}$. 

For each $t$, we cluster the set of $K_t^{(i)},i=1,...,N$, where $N$ is the size of the training set. The output is a set of clusters $C_t = \{c_j\}_{j=1}^{P_t}$, where $P_t$ is the number of clusters for $t$, and correspond to distinct part configurations 

\begin{equation}
  C_t = \textit{cluster}\left(\{K_t^{(i)}\}_{i=1}^N\right).
\eqlabel{cluster}
\end{equation}

We use a greedy clustering algorithm, similar to \cite{armlets2013}. Examples are processed in a random order. An example is added to an existing cluster if its distance to the center is less than $\epsilon$, otherwise it starts a new cluster. The distance of two examples is defined as the euclidean distance of their normalized keypoint distributions. For each cluster $c \in C_t$, we collect the $M$ closest cluster members to its center. Those form the set of positive examples that represent the cluster. From now on, we describe a part by its body part type $t$  and its cluster index $j$, with $c_j \in C_t$ while $S_{t,j}$ represents the set of positive examples for part $(t,j)$.

\figref{clusters} (left) shows examples of clusters as produced by our clustering algorithm with $\epsilon = 1$ and $M=100$. We show 4 examples for each cluster example. We use the PASCAL VOC 2012 train set, along with keypoint annotations as provided by \cite{BourdevPoseletsECCV10}, to design and train the part detectors. In total we obtain 30 parts, 13 for head, 11 for torso and 6 for legs.

\begin{figure}[t]
\begin{center}
  \includegraphics[width=0.8\linewidth]{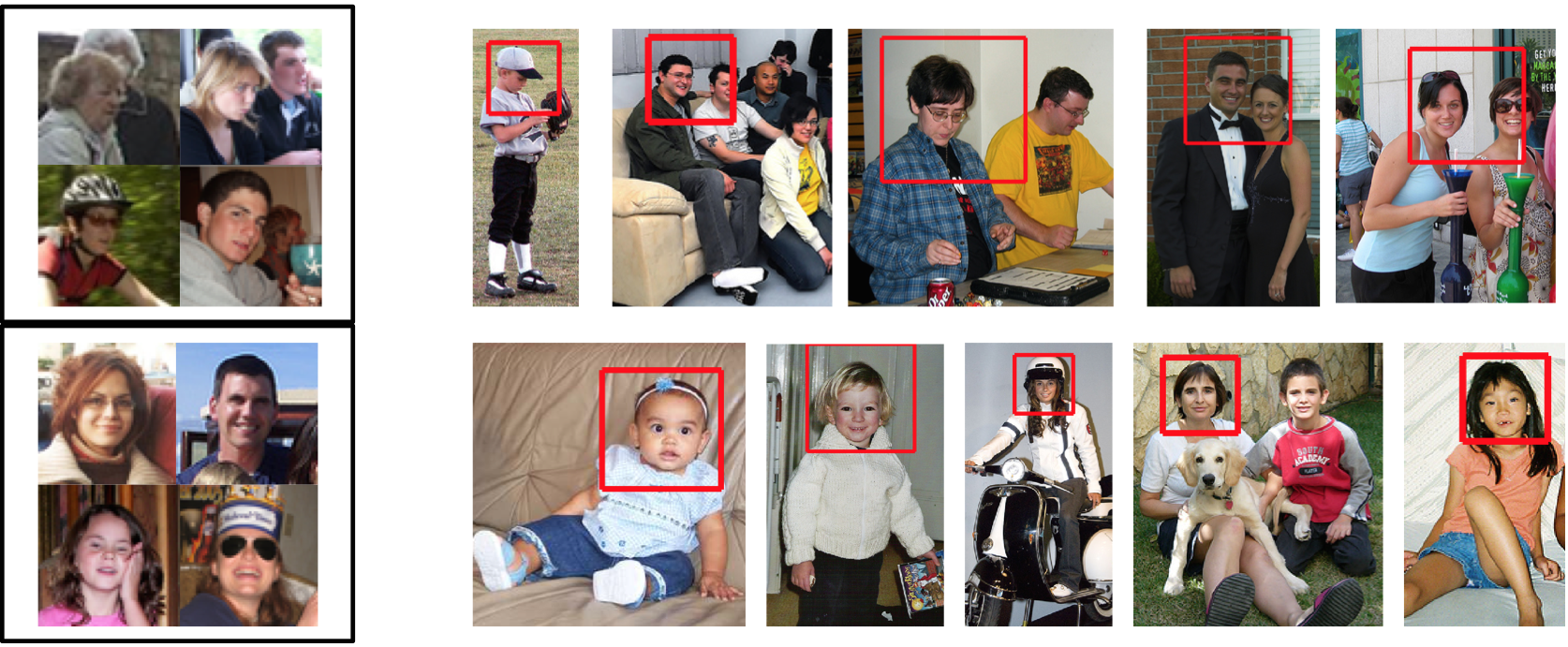}
  \includegraphics[width=0.8\linewidth]{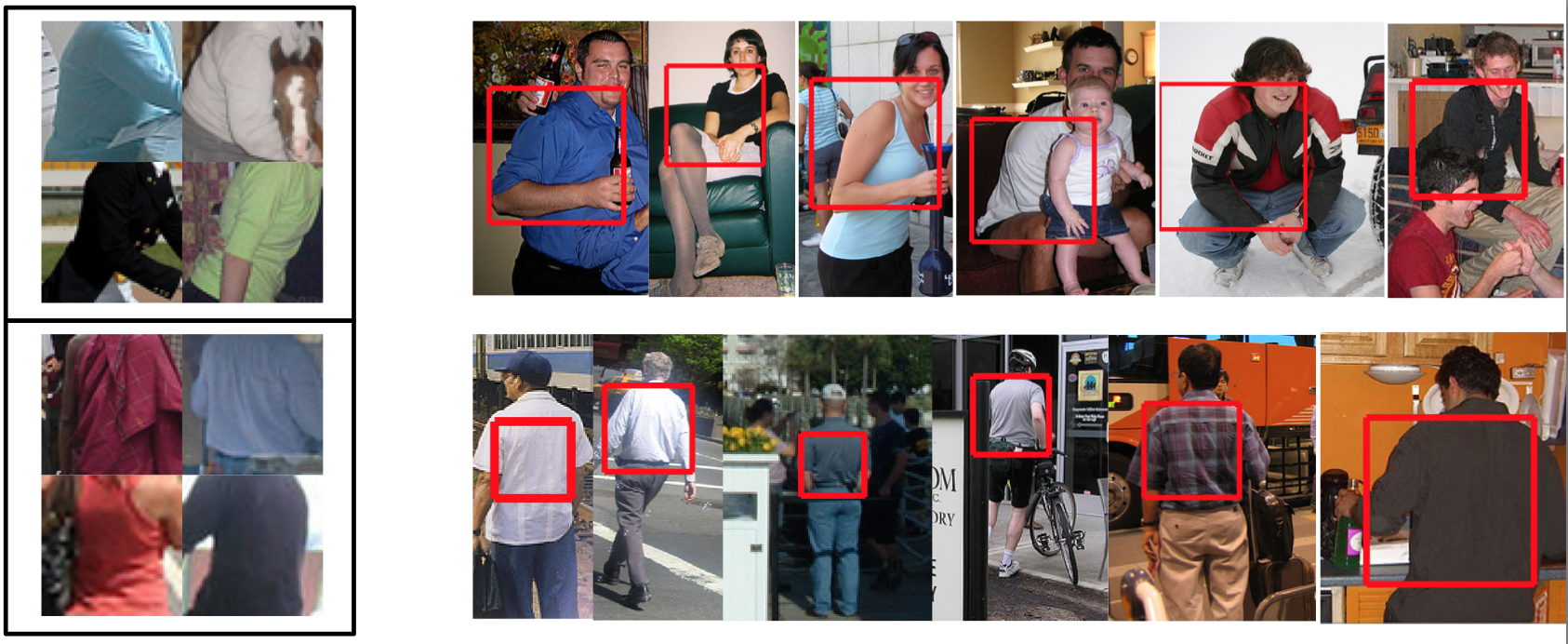}
  \includegraphics[width=0.8\linewidth]{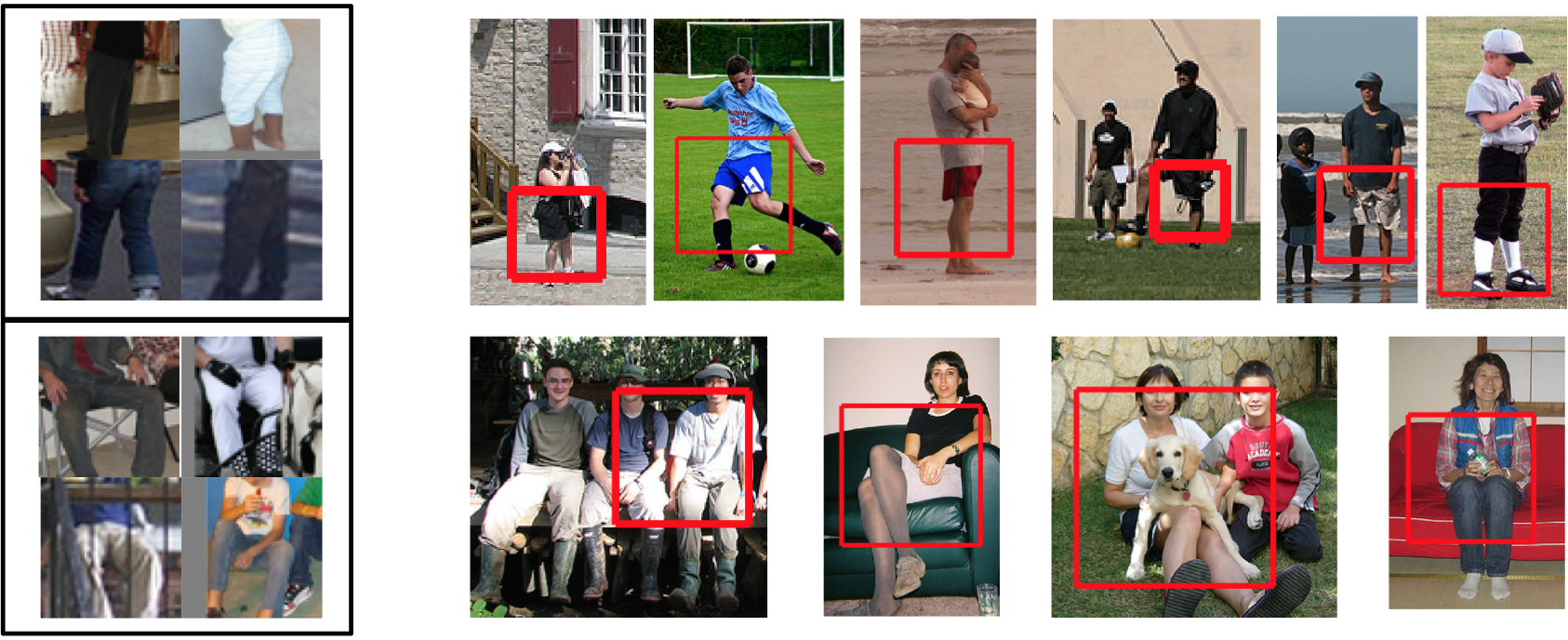}
\end{center}
\caption{Examples of clusters for the three body areas, head, torso and legs (left) and their top few detections on PASCAL VOC val 2009 (right). The first two rows correspond to cluster examples for head, the following two for torso and the last two for legs.}
   \figlabel{clusters}
\end{figure}

\subsubsection{Learning part models}

For each part $(t,j)$, we define the part model to be the vector of weights ${\bf w}_{t,j}$ which when convolved with a feature pyramid gives stronger activations near the ground-truth location and scale (right most part of \figref{DeepParts}).

One could view the whole pipeline shown in \figref{DeepParts} as a fully convolutional model and thus one could train it end-to-end, optimizing the weights of the CNN for the pool5 feature extraction and the weights of the part models jointly. We choose to simplify the problem by decoupling it. We use the publicly available ImageNet weights of the CNN \cite{girshick2014rcnn} to extract pool5 feature pyramids. Subsequently, we train linear SVMs for the part models. For each part $(t,j)$ we train a linear SVM with positives from $S_{t,j}$ to obtain model weights ${\bf w}_{t,j} \in \mathbb{R}^{8\times8\times256}$. We use hard negative mining from images of no people to train the model.

\figref{clusters} (right) shows the top few detections of a subset of parts on PASCAL VOC val 2009 set. Each row shows activations of a different part, which is displayed at the left part of the same row.

\paragraph{Evalutation of part models.} We quantify the performance of our part detectors by computing the average precision (AP) - similar to object detection PASCAL VOC - on val 2009. For every image, we detect part activations at all scales and locations which we non-maximum suppress with a threshold of 0.3 across all parts of the same type. Since there are available keypoint annotations on the val set, we are able to construct ground-truth part boxes. A detection is marked as positive if the intersection-over-union with a ground-truth part box is more than $\sigma$. In PASCAL VOC, $\sigma$ is set to 0.5. However, this threshold is rather strict for small objects, such as our parts. We report AP for various values of $\sigma$ for a fair assessment of the quality of our parts. \tableref{ap_parts} shows the results. 

\begin{table}[t!]
\centering
\renewcommand{\arraystretch}{1.2}
\renewcommand{\tabcolsep}{1.2mm}
\resizebox{0.8\linewidth}{!}{
\begin{tabular}{@{}l|r*{5}{c}|cc@{}}
AP (\%)  & $\sigma=0.2$ & $\sigma=0.3$ & $\sigma=0.4$ & $\sigma=0.5$ \\
\hline
Head & 55.2 & 51.8 & 45.2 & 31.6 \\
Torso & 42.1 & 36.3 & 23.6 & 9.4\\
Legs & 34.9 & 27.9 & 20.0 & 10.4
\end{tabular}
}
\vspace{0.1em}
\caption{AP for each part type on PASCAL VOC val 2009. We evaluate the part activations and measure AP for different thresholds of intersection-over-union.}
\tablelabel{ap_parts}
\vspace{-0.5em}
\end{table}

\paragraph{Mapping parts to instances.}
Since our part models operate independently, we need to group part activations and link them to an instance in question. Given a candidate region $box$ in an image $I$, for each part $t$ we keep the highest scoring part within $box$
\begin{equation}
  j^* = \argmax_j \max_{(x,y) \in box} \mathbf{w}_{t,j} \ast F_{(x,y)}(I),
\end{equation}
where $F_{(x,y)}(I)$ is the point in feature pyramid for $I$ corresponding to the image coordinates $(x,y)$. This results in three parts being associated with each $box$, as shown in \figref{fig1}. A part is considered absent if the score of the part activation is below a threshold, here the threshold is set to -0.1.

In the case when an oracle gives ground-truth bounding boxes at test time, one can refine the search of parts even further. If $box$ is the oracle box in question, we retrieve the $k$ nearest neighbor instances $i=\{i_1,...,i_k\}$from the training set based on the $L_2$-norm of their pool5 feature maps $F(\cdot)$, \ie $\frac{F(box)^T F(box_{i_j})}{||F(box)|| \cdot ||F(box_{i_j})||}$. If $K_{i_j}$ are the keypoints for the nearest examples, we consider the average keypoint locations $K_{box} = \frac{1}{K} \sum_{j=1}^k K_{i_j}$ to be an estimate of the keypoints for the test instance $box$. Based on $K_{box}$ we can reduce the regions of interest for each part within $box$ by only searching for them in the corresponding estimates of the body parts. 

\section{Part-based Classification}
\seclabel{tasks}

In this section we investigate the role of parts for fine-grained classification tasks.
We focus on the tasks of action classification (e.g. running, reading, etc.) and attribute classification (e.g. male, wears hat, etc.).
\figref{FineGrained} schematically outlines our approach at test time.
We start with the part activations mapped to an instance and forward propagate the corresponding part and instance boxes through a CNN.
The output is a fc7 feature vector for each part as well as the whole instance.
We concatenate the feature vectors and classify the example with a linear SVM, which predicts the confidence for each class (action or attribute).

\subsection{System variations}
For each task, we consider four variants of our approach in order to understand which design factors are important.

\paragraph{No parts.}
This approach is our baseline and does not use part detectors.
Instead, each instance is classified according to the fc7 feature vector computed from the instance bounding box.
The CNN used for this system is fine-tuned from an ImageNet initialization, as in \cite{girshick2014rcnn}, on jittered instance bounding boxes.

\paragraph{Instance fine-tuning.}
This method uses our part detectors.
Each instance is classified based on concatenated fc7 feature vectors from the instance and all three parts.
The CNN used for this system is fine-tuned on instances, just as in the ``no parts'' system.
We note that because some instances are occluded, and due to jittering, training samples may resemble parts, though typically only the head and torso (since occlusion tends to happen from the torso down).

\paragraph{Joint fine-tuning.}
This method also uses our part detectors and concatenated fc7 feature vectors.
However, unlike the previous two methods we fine-tune the CNN jointly using instance and part boxes from each training sample.
During fine-tuning the network can be seen as a four-stream CNN, with one stream for each bounding box.
Importantly, we tie weights between the streams so that the number of CNN parameters is the same in all system variants.
This design explicitly forces the CNN to see each part box during fine-tuning.

\paragraph{3-way split.}
To test the importance of our part detectors, we employ a baseline that vertically splits the instance bounding box into three (top, middle, and bottom) in order to simulate crude part detectors.
This variation uses a CNN fine-tuned on instances.

\begin{figure*}[t]
\begin{center}
  \includegraphics[width=0.9\linewidth]{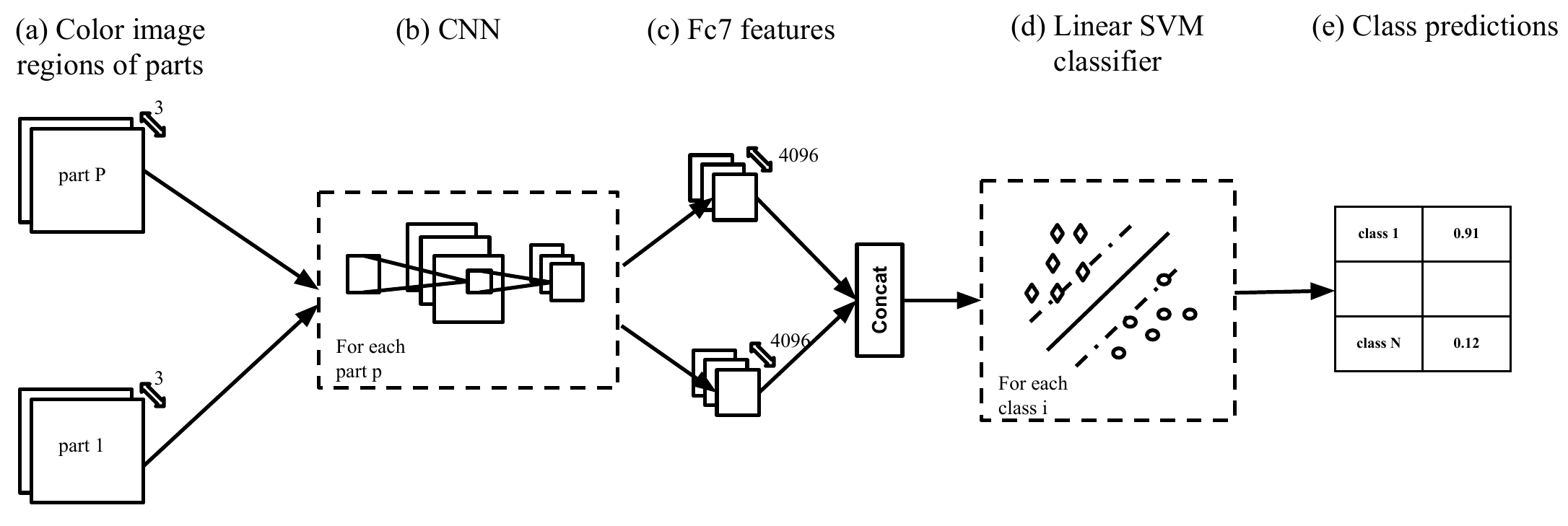}
\end{center}
   \caption{Schematic overview of our approach for fine grained classification using parts. (a) We consider regions of part activations. (b) Each part is forward propagated through a CNN. (c) The output is the fc7 feature vector for each input. (d) The features are concatenated and fed into linear SVM classifiers. (e) The classifiers produce scores for each class.}
   \figlabel{FineGrained}
\end{figure*} 

\subsection{Action Classification}

We focus on the problem of action classification as defined by the PASCAL VOC action challenge. The task involves predicting actions from a set of predefined action categories.

\paragraph{Learning details.} 
 
We train all networks with backpropagation using Caffe \cite{jia2014caffe}, starting from the ImageNet weights, similar to the fine-tuning procedure introduced in \cite{girshick2014rcnn}.
A small learning rate of $10^{-5}$ and a dropout ratio of 50\% were used.
During training, and at test time, if a part is absent from an instance then we use a box filled with the ImageNet mean image values (\ie, all zeros after mean subtraction).
Subsequently, we train linear SVMs, one for each action, on the concatenated fc7 feature vectors.

\paragraph{Context.}
In order to make the most of the context in the image, we rescore our predictions by using the output of R-CNN \cite{girshick2014rcnn} for the 20 PASCAL VOC object categories and the presence of other people performing actions. We train a linear SVM on the action score of the test instance, the maximum scores of other instances (if any) and the object scores, to obtain a final prediction. 
Context rescoring is used for all system variations on the \emph{test} set.

\paragraph{Results.}
\tableref{action} shows the result of our approach on the PASCAL VOC 2012 \emph{test} set. These results are in the standard setting, where an oracle gives ground-truth person bounds at test time. We conduct experiments using two different network architectures: a 8-layer CNN as defined in \cite{krizhevsky2012imagenet}, and a 16-layer as defined in \cite{vgg}. \emph{Ours (no parts)} is the baseline approach, with no parts. \emph{Ours} is our full approach when we include the parts. For the 8-layer network, we use the CNN trained on instances, while for the 16-layer network we use the CNN trained jointly on instances and their parts based on results on the \emph{val} set (\tableref{action_dtc}). For our final system, we also present results when we add features extracted from the whole image, using a 16-layer network trained on ImageNet-1k (\emph{Ours (w/ image features)}). 
We show results as reported by action poselets \cite{MajiActionCVPR11}, a part-based approach, using action specific poselets with HOG features, Oquab \etal \cite{Oquab14},  Hoai \cite{Hoai-BMVC14} and Simonyan and Zisserman \cite{vgg}, three CNN-based approaches on the task. The best performing method by \cite{vgg} uses a 16- and 19-layer network. Their 16-layer network is equivalent to \emph{Ours (no parts)} with 16 layers, thus the additional boost in performance comes from the 19-layer network. This is not surprising, since deeper networks perform better, as is also evident from our experiments. From the comparison with the baseline, we conclude that parts improve the performance. For the 8-layer CNN, parts contribute 3\% of mAP, with the biggest improvement coming from \emph{Phoning}, \emph{Reading} and \emph{Taking Photo}. For the 16-layer CNN, the improvement from parts is smaller, 1.7 \% of mAP, and the actions benefited the most are  \emph{Reading}, \emph{Taking Photo} and \emph{Using Computer}. The image features capture cues from the scene and give an additional boost to our final performance.

\tableref{action_dtc} shows results on the PASCAL VOC action \emph{val} set for a variety of different implementations of our approach. \emph{Ours (no parts)} is the baseline approach, with no parts, while \emph{Ours (3-way split)} uses as parts the three horizontal splits comprising the instance box. \emph{Ours (joint fine-tuning)} shows the results when using a CNN fine-tuned jointly on instances and parts, while \emph{Ours (instance fine-tuning)} shows our approach when using a CNN fine-tuned on instances only.
We note that all variations that use parts significantly outperform the \emph{no-parts} system.

We also show results of our best system when ground-truth information is \emph{not} available at test time \emph{Ours (R-CNN bbox)}. In place of oracle boxes we use R-CNN detections for person. For evaluation purposes, we associate a R-CNN detection to a ground-truth instance as following: we pick the highest scoring detection for person that overlaps more than 0.5 with the ground truth. Another option would be to define object categories as ``person+action" and then just follow the standard detection AP protocol. However, this is not possible because not all people are marked in the dataset (this is true for the attribute dataset as well). We report numbers on the val action dataset. We observe a drop in performance, as expected due to the imperfect person detector, but our method still works reasonably well under those circumstances.
\figref{action_res} shows the top few predictions on the test set. Each block corresponds to a different action.

\begin{table*}[t!]
\centering
\renewcommand{\arraystretch}{1.2}
\renewcommand{\tabcolsep}{1.2mm}
\resizebox{\linewidth}{!}{
\begin{tabular}{@{}l|c|r*{9}{c}|cc@{}}
AP (\%)  & CNN layers & Jumping  & Phoning & Playing Instrument & Reading & Riding Bike & Riding Horse & Running & Taking Photo & Using Computer & Walking & mAP \\
\hline
Action Poselets \cite{MajiActionCVPR11}  & - & 59.3 & 32.4 & 45.4 & 27.5 & 84.5 & 88.3 & 77.2 & 31.2 & 47.4 & 58.2 &  55.1\\
Oquab \etal \cite{Oquab14}                        & 8 & 74.8 & 46.0 & 75.6 & 45.3 & 93.5 & 95.0 & 86.5 & 49.3 & 66.7 & 69.5 & 70.2 \\
Hoai \cite{Hoai-BMVC14}                           & 8 & 82.3 & 52.9 & 84.3 & 53.6 & 95.6 & 96.1 & 89.7 & 60.4 & 76.0  & \bf{72.9} & 76.3 \\
Simonyan \& Zisserman \cite{vgg}              & 16 \& 19 & \bf{89.3} & \bf{71.3} & \bf{94.7} & \bf{71.3} & \bf{97.1}& \bf{98.2} & \bf{90.2} & 73.3 & \bf{88.5} & 66.4 & \bf{84.0} \\
\hline
Ours (no parts)                             & 8 & 76.2 & 47.4 & 77.5 & 42.2 & 94.9 & 94.3 & 87.0 & 52.9 & 66.5 & 66.5 & 70.5 \\
Ours                                               & 8 & 77.9 & 54.5 & 79.8 & 48.9 & 95.3 & 95.0 & 86.9 & 61.0 & 68.9 & 67.3 & 73.6 \\
Ours (no parts)                           & 16 & 84.7 & 62.5 & 86.6 & 59.0 & 95.9 & 96.1 & 88.7 & 69.5 & 77.2 & 70.2 & 79.0 \\
Ours                                             & 16 & 83.7 & 63.3 & 87.8 & 64.2 & 96.0 & 96.7 & 88.9 & 75.2 & 80.0 & 71.5 & 80.7 \\
Ours (w/ image features)            & 16 &  84.7 & 67.8 & 91.0 & 66.6 & 96.6 & 97.2 & \bf{90.2} & \bf{76.0} & 83.4 & 71.6 & 82.6
\end{tabular}
}
\vspace{0.1em}
\caption{AP on the PASCAL VOC 2012 Actions test set.  The first three rows show results of two other methods. Action Poselets \cite{MajiActionCVPR11} is a part based approach using HOG features, while Oquab \etal \cite{Oquab14}, Hoai \cite{Hoai-BMVC14} and Simonyan \& Zisserman \cite{vgg} are CNN based approaches. \emph{Ours (no parts)} is the baseline approach of our method, when only the ground truth box is considered, while \emph{Ours} is the full approach, including parts. All approaches use ground truth boxes at test time.}
\tablelabel{action}
\vspace{-0.5em}
\end{table*}

\begin{table*}[t]
\centering
\renewcommand{\arraystretch}{1.2}
\renewcommand{\tabcolsep}{1.2mm}
\resizebox{\linewidth}{!}{
\begin{tabular}{@{}l|c|r*{9}{c}|cc@{}}
AP (\%)  & CNN layers & Jumping  & Phoning & Playing Instrument & Reading & Riding Bike & Riding Horse & Running & Taking Photo & Using Computer & Walking & mAP \\
\hline
Ours (no parts)                             & 8 & 76.5 & 44.7 & 75.0 & 43.3 & 90.0 & 91.6 & 79.2 & 53.5 & 66.5 & 61.4 & 68.2 \\
Ours (3-way split)                          & 8 & 79.0 & 44.4 & 77.8 & 46.9 & 91.5 & 93.1 & 83.1 & 59.3 & 67.3 & 64.4 & 70.7 \\
Ours (instance fine-tuning)           & 8 & 77.2 & 48.4 & 79.0 & 49.5 & 92.4 & 93.8 & 80.9 & 60.4 & 68.9 & 64.0 & 71.5  \\
Ours (joint fine-tuning)                  & 8 & 75.2 & 49.5 & 79.5 & 50.2 & 93.7 & 93.6 & 81.5 & 58.6 & 64.6 & 63.6 & 71.0  \\
Ours (no parts) 			     & 16 & \bf{85.4} & 58.6 & 84.6 & 60.9 & 94.4 & 96.6 & \bf{86.6} & 68.7 & 74.9 & 67.3 & 77.8 \\
Ours (instance fine-tuning)	     & 16 &  85.1 & 60.2 & \bf{86.6} & 63.1 & 95.6 & 97.4 & 86.4 & 71.0 & 77.6 & 68.3 & 79.1 \\
Ours (joint fine-tuning)                & 16 & 84.5 & \bf{61.2} & 88.4 & \bf{66.7} & \bf{96.1} & \bf{98.3} & 85.7 & \bf{74.7} & \bf{79.5} & \bf{69.1} & \bf{80.4} \\
\hline
Ours (R-CNN bbox)                     & 8 & 67.8  & 46.6 & 76.9 & 47.3 & 85.9 & 81.4 & 71.5 & 53.1 & 61.2 & 53.9 & 64.6 \\
Ours (R-CNN bbox)                      & 16 & 79.4 & 63.3 & 86.1 & 64.4 & 93.2 & 91.9 & 80.2 & 71.2 & 77.4 & 63.4 & 77.0 \\
\end{tabular}
}
\vspace{0.1em}
\caption{AP on the PASCAL VOC 2012 Actions val set of our approach. \emph{Ours (no parts)} is our approach without parts. \emph{Ours (3-way split)} is our approach when parts are defined as the three horizontal splits comprising an instance box. \emph{Ours (joint fine-tuning)} uses a CNN fine-tuned jointly on the instances and the parts, while \emph{Ours (instance fine-tuning)} uses a single CNN fine-tuned just on the instance box. All the above variations of our approach use ground truth information at test time as the object bound. \emph{Ours (R-CNN bbox)} uses R-CNN detections for person.}
\tablelabel{action_dtc}
\vspace{-0.5em}
\end{table*}

\begin{figure*}[htb!]
\begin{center}
  \includegraphics[width=0.81\linewidth]{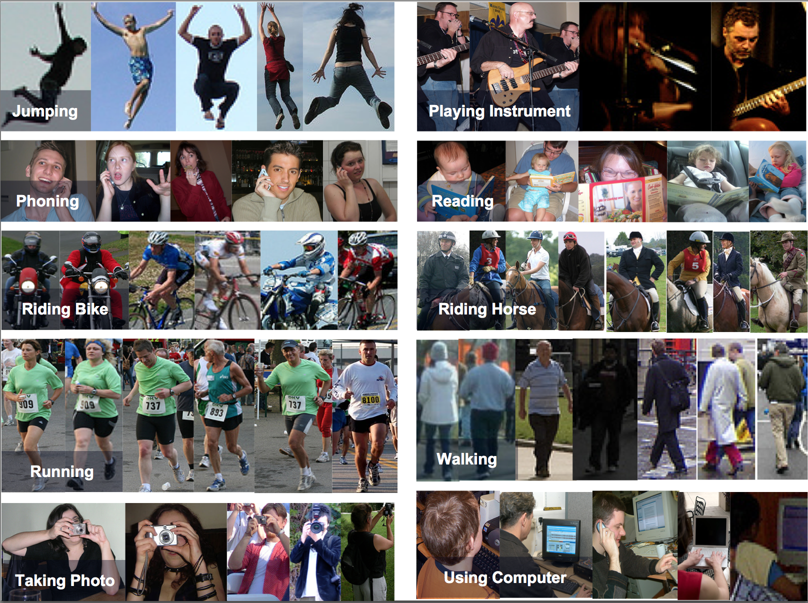}
\end{center}
   \caption{Top action predictions on the test set. Different blocks correspond to different actions.}
   \figlabel{action_res}
\end{figure*} 

\subsection{Attribute Classification}

\begin{table*}[htb!]
\centering
\renewcommand{\arraystretch}{1.2}
\renewcommand{\tabcolsep}{1.2mm}
\resizebox{\linewidth}{!}{
\begin{tabular}{@{}l|c|r*{8}{c}|cc@{}}
AP (\%)  & CNN layers & Is Male & Has Long Hair & Has Glasses & Has Hat & Has T-Shirt & Has Long Sleeves & Has Shorts & Has Jeans & Has Long Pants & mAP \\
\hline
PANDA \cite{panda}                     & 5 & 91.7 & 82.7 & 70.0 & 74.2 & 49.8 & 86.0 & 79.1 & 81.0 & 96.4 & 79.0 \\
Ours (no parts)                             & 8 & 87.5 & 80.4 & 43.3 & 77.0 & 61.5 & 86.4 & 88.5 & 88.7 & 98.2 & 79.1\\
Ours (3-way split)                         & 8 & 89.3 & 82.2 & 51.2 & 84.0 & 60.1 & 87.4 & 88.3 & 89.2 & 98.2 & 81.1 \\
Ours (instance fine-tuning)          & 8 & 89.9 & 83.5 & 60.5 & 85.2 & 64.3 & 89.0 & 88.6 & 89.1 & 98.2 & 83.1 \\
Ours (joint fine-tuning)                 & 8 & 91.7 & 86.3 & 72.5 & 89.9 & 69.0 & 90.1 & 88.5 & 88.3 & 98.1 & 86.0 \\
Ours (no parts)   		          & 16 & 93.4 & 88.7 & 72.5 & 91.9 & 72.1 & 94.1 & 92.3 & 91.9 & 98.8 & 88.4 \\
Ours (instance fine-tuning)         & 16 & \bf{93.8} & 89.8 & 76.2 & 92.9 & \bf{73.3} & \bf{94.4} & 92.3 & 91.8 & 98.7 & 89.3 \\
Ours (joint fine-tuning)               & 16 & 92.9 & \bf{90.1} & \bf{77.7} & \bf{93.6} & 72.6 & 93.2 & \bf{93.9} & \bf{92.1} & \bf{98.8} & \bf{89.5} \\
\hline
Ours (R-CNN bbox)                   & 8 & 84.1 & 77.9 & 62.7 & 84.5 & 66.8 & 84.7 & 80.7 & 79.2 & 91.9 & 79.2 \\
Ours (R-CNN bbox)                  & 16 & 90.1 & 85.2 & 70.2 & 89.8 & 63.2 & 89.7 & 83.4 & 84.8 & 96.3 & 83.6 \\
\end{tabular}
}
\vspace{0.1em}
\caption{AP on the test set of the Berkeley Attributes of People Dataset. All approaches on the top use ground truth boxes for evaluation. \emph{Ours (no parts)} is the baseline approach with no parts. \emph{Ours (3-way split)} is a variant of our approach, where parts are defined as the three horizontal splits comprising an instance box. \emph{Ours (instance fine-tuning)} uses a CNN fine-tuned on instance boxes, while \emph{Ours (joint fine-tuning)} uses a CNN fine-tuned jointly on instances and parts. We also show the effectiveness of our approach \emph{Ours (R-CNN bbox)}, when no ground truth boxes are given at test time.}
\tablelabel{attributes}
\vspace{-0.5em}
\end{table*}

We focus on the problem of attribute classification, as defined by \cite{BourdevAttributesICCV11}. There are 9 different categories of attributes, such as \emph{Is Male}, \emph{Has Long Hair}, and the task involves predicting attributes, given the location of the people.  Our approach is shown in \figref{FineGrained}. We use the Berkeley Attributes of People Dataset as proposed by \cite{BourdevAttributesICCV11}. 

\paragraph{Learning details.} Similar to the task of action classification, we separately learn the parameters of the CNN and the linear SVM. Again, we fine-tune a CNN for the task in question with the difference that the softmax layer is replaced by a cross entropy layer (sum of logistic regressions). 

\paragraph{Results.}

\tableref{attributes} shows AP on the \emph{test} set. We show results of our approach with and without parts, as well as results as reported by Zhang \etal \cite{panda}, the state-of-the-art on the task, on the same test set. With an 8-layer network, parts improve the performance of all categories, indicating their impact on attribute classification. Also, a network jointly fine-tuned on instances and parts seems to work significantly better than a CNN trained solely on the instance boxes.
In the case of a 16-layer network, joint fine-tuning and instance fine-tuning seem to work equally well. The gain in performance from adding parts is less significant in this case. This might be because of the already high performance achieved by the holistic network.
Interestingly, our 8-layer holistic approach matches the current state-of-the-art on this task, PANDA \cite{panda} showcasing the importance of deeper networks and good initialization.

\tableref{attributes} also shows the effectiveness of our best model, namely the jointly fine-tuned 16-layer CNN, when we use R-CNN detections instead of ground truth boxes on the Berkeley Attributes of People test set.
\figref{attr_res} shows the top few predictions on the test set. Each block corresponds to a different attribute. \figref{attr_errs} shows top errors for two of our lowest performing attribute classes. 

\begin{figure}
\centering  
\includegraphics[width=0.7\linewidth]{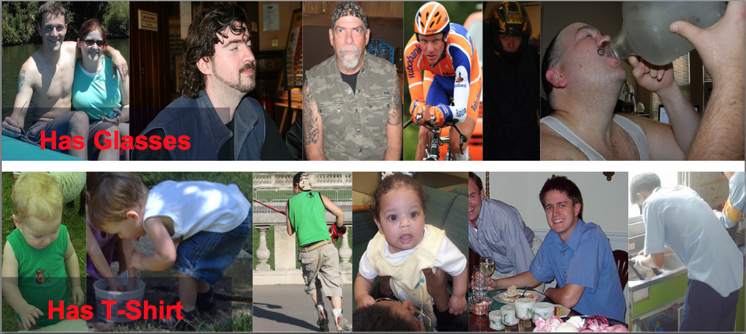}
\caption{Top errors of classification for two of the attribute categories, \emph{Has Glasses} (top) and \emph{Has T-Shirt} (bottom).}
\figlabel{attr_errs}
\end{figure}

\begin{figure*}[htb!]
\centering  
  \includegraphics[width=0.81\linewidth]{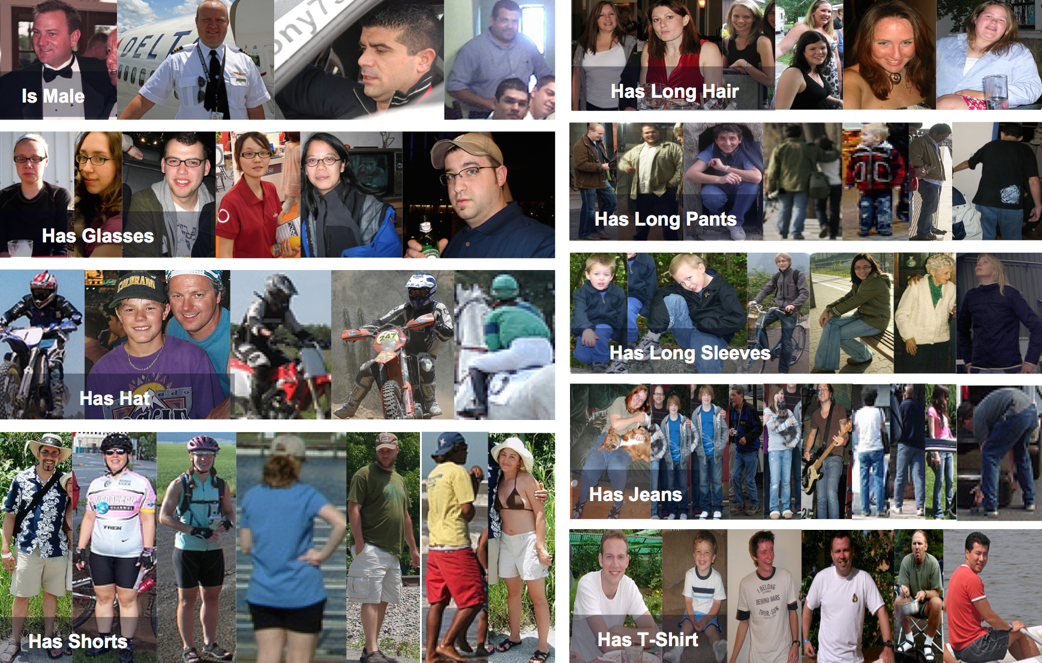}
   \caption{Top attribute predictions on the test set. Each block corresponds to a different attribute}
\figlabel{attr_res}
\end{figure*}



\newpage
{\small
\bibliographystyle{ieee}
\bibliography{refs}
}

\end{document}